\documentclass{article}

\usepackage{arxiv}

\usepackage[utf8]{inputenc} 
\usepackage[T1]{fontenc}    
\usepackage{hyperref}       
\usepackage{url}            
\usepackage{booktabs}       
\usepackage{amsfonts}       
\usepackage{nicefrac}       
\usepackage{microtype}      
\usepackage{lipsum}
\usepackage{bm}
\usepackage{amsmath}
\usepackage{color}
\usepackage{graphicx}
\usepackage{siunitx}
\usepackage{algorithm,algpseudocode}

\title{Generating valid Euclidean distance matrices}

\author{
	Moritz Hoffmann\\
	Department of Mathematics and Computer Science\\
	Freie Universität Berlin\\
	Arnimallee 6, 14195 Berlin, Germany
	\And
	Frank Noé\thanks{Correspondence to: \texttt{frank.noe@fu-berlin.de}}  \\
	Department of Mathematics and Computer Science\\
	Freie Universität Berlin\\
	Arnimallee 6, 14195 Berlin, Germany
}

\begin{document}
	\maketitle
	
	\begin{abstract}
		Generating point clouds, e.g., molecular structures, in arbitrary rotations, translations, and enumerations remains a challenging task.
		Meanwhile, neural networks utilizing symmetry invariant layers have been shown to be able to optimize their training objective in a data-efficient way.
		In this spirit, we present an architecture which allows to produce
		valid Euclidean distance matrices, which by construction are already invariant
		under rotation and translation of the described object.
		Motivated by the goal to generate molecular structures in Cartesian space, we use this architecture to construct a Wasserstein GAN utilizing a permutation invariant critic network.
		This makes it possible to generate molecular structures in a one-shot fashion by producing Euclidean distance matrices which have a three-dimensional embedding.
	\end{abstract}

	\section{Introduction}
	
	Recently there has been great interest in deep learning based on graph
	structures \cite{defferrard2016convolutional,kipf2016semi,gilmer2017neural} and point clouds \cite{qi2017pointnet,li2018pointcnn,yang2019pointflow}.
	A prominent application example is that of molecules, for which both
	inference based on the chemical compound, i.e., the molecular graph
	structure \cite{kearnes2016molecular,janet2017predicting,WinterEtAl_Translation_Chemistry2019},
	and based on the geometry, i.e. the positions of atoms in 3D space
	\cite{BehlerParrinello_PRL07_NeuralNetwork,RuppEtAl_PRL12_QML,schutt2017schnet,SmithIsayevRoitberg_ChemSci17_ANI}
	are active areas of research.
	
	A particularly interesting branch of machine learning for molecules
	is the reverse problem of generating molecular structures, as it opens
	the door for designing molecules, e.g., obtain new materials~\cite{sanchez2018inverse,barnes2018machine,elton2018applying,li2018tuning},
	design or discover pharmacological molecules such as inhibitors or antibodies~\cite{popova2018deep,griffen2018can},
	optimize biotechnological processes~\cite{guimaraes2017objective}. While
	this area of research has exploded in the past few years, the vast
	body of work has been done on the generation of new molecular compounds,
	i.e. the search for new molecular graphs, based on string encodings
	of that graph structure or other representations \cite{GomezBombarelli_ACSCentral_AutomaticDesignVAE,WinterEtAl_MultiobjectiveOptimization}.
	On the other hand, exploring the geometry space of the individual chemical compound
	is equally important, as the molecular geometries and their probabilities
	determine all equilibrium properties, such as binding affinity, solubility
	etc. Sampling different geometric structures is, however, still largely
	left to molecular dynamics (MD) simulation that suffers from the rare
	event sampling problem, although recently machine learning has been
	used to speed up MD simulation \cite{RibeiroTiwary_JCP18_RAVE,BonatiZhangParrinello_PNAS19_VariationalNN,ZhangYanNoe_ChemRxiv19_TALOS,PlattnerEtAl_NatChem17_BarBar,DoerrDeFabritiis_JCTC14_OnTheFly} or to perform sampling
	of the equilibrium distribution directly, without MD \cite{NoeEtAl_19_BoltzmannGenerators}.
	All of these techniques only sample one single chemical compound in geometry space.
	
	Here we explore---to our best knowledge for the first time in depth---the simultaneous generation of chemical compounds and geometries. The only related work we are aware of~\cite{gebauer2018generating,gebauer2019symmetry} demonstrates the generation of chemical compounds, placing
	atom by atom with an autoregressive model. It was shown that the model can recover compounds contained in the QM9 database of small molecules~\cite{ruddigkeit2012enumeration,ramakrishnan2014quantum} when trained on a subset, but different configurations of the same molecule were not analyzed.
	
	While autoregressive models seem to work well in the case of small ($< 9$ heavy atoms) molecules like the ones in the QM9 database,
	they can be tricky for larger structures as the probability to completely form complex structures, such as rings, decays with the number of involved steps.
	
	To avoid this limitation, in our method we investigate in one shot models for point clouds which have no absolute orientation, i.e.,
	the point cloud structure is considered to be the same independent of its rotation, translation, and of the permutation of points.
	
	A natural representation, which is independent of rotation and translation is the Euclidean distance matrix,
	which is the matrix of all squared pairwise Euclidean distances. Furthermore, Euclidean distance matrices are useful determinants of valid molecular structures.
	
	While a symmetric and non-negative matrix with a zero diagonal can easily be parameterized
	by, e.g., a neural network, it is not immediately clear that this matrix indeed belongs to a set
	of $n$ points in Euclidean space and even then, that this space has the right dimension.
	
	Here we develop a new method to parameterize and generate valid Euclidean distance matrices
	without placing coordinates directly, hereby taking away a lot of the ambiguity.
	
	We furthermore propose a Wasserstein GAN architecture for learning distributions of pointclouds, e.g., molecular
	structures invariant to rotation, translation, and permutation. To this end the data distribution
	as well as the generator distribution are represented in terms of Euclidean distance matrices.
	
	In summary, our main contributions are as follows:
	\begin{itemize}
		\item We introduce a new method of training neural networks so that their output are Euclidean distance matrices with a predefined embedding dimension.
		\item We propose a GAN architecture, which can learn distributions of Euclidean distance matrices, while treating the structures described by the distance matrices as set, i.e., invariant under their permutations.
		\item We apply the proposed architecture to a set of $\mathrm{C}_7\mathrm{O}_2\mathrm{H}_{10}$ isomers contained in the QM9 database and show that it can recover parts of the training set as well as generalize out of it.
	\end{itemize}

	Code is available at \href{https://github.com/noegroup/EDMnets}{https://github.com/noegroup/EDMnets}.
	
	\section{Generating Euclidean distance matrices\label{sec:edm}}
	
	We describe a way to generate Euclidean distance matrices
	$D\in\mathbb{EDM}^{n}\subset\mathbb{R}^{n\times n}$ without placing coordinates in
	Cartesian space. This means in particular that the parameterized output
	is invariant to translation and rotation.
	
	A matrix $D$ is in $\mathbb{EDM}^{n}$ by definition if there exist
	points $\bm{x}_{1},\ldots,\bm{x}_{n}\in\mathbb{R}^{d}$ such
	that $D_{ij}=\|\bm{x}_{i}-\bm{x}_{j}\|_{2}^{2}$ for
	all $i,j=1,\ldots,n$. Such a matrix $D$ defines a structure in Euclidean
	space $\mathbb{R}^{d}$ up to a combination of translation, rotation,
	and mirroring. The smallest integer $d>0$ for which a set of $n$ points
	in $\mathbb{R}^d$ exists that reproduces the matrix $D$ is called the
	embedding dimension.
	
	The general idea of the generation process is to produce a hollow (i.e., zeros on the diagonal)
	symmetric matrix $\tilde{D}$ and then weakly enforce $\tilde{D}\in\mathbb{EDM}^{n}$
	through a term in the loss. It can be shown that
	\begin{equation}
	\tilde{D}\in\mathbb{EDM}^{n}\Leftrightarrow-\frac{1}{2}J\tilde{D}J\,\text{positive semi-definite},\label{eq:edmcondition}
	\end{equation}
	where $J=\mathbb{I}-\frac{1}{n}\bm{1}\bm{1}^{\top}$
	and $\bm{1}=(1,\ldots,1)^{\top}\in\mathbb{R}^{n}$ \cite{schoenberg1935remarks,gower1982euclidean}. However trying to use this relationship directly in the
	context of deep learning by parameterizing the matrix $\tilde{D}$
	poses a problem, as the set of EDMs forms a cone~(\cite{dattorro2010convex}) and not a vector
	space, which is the underlying assumption of the standard optimizers
	in common deep learning frameworks. One can either turn to
	optimization techniques on Riemannian manifolds~(\cite{zhang2016riemannian})
	or find a reparameterization in which the
	network's output behaves like a vector space and that can be transformed into an EDM.
	
	Here, we leverage a connection between EDMs and positive semi-definite
	matrices \cite{alfakih1999solving,krislock2012euclidean} in order to parameterize the problem in a space that behaves like a vector space. In particular,
	for $D\in\mathbb{EDM}^{n}$ by definition there exist points
	$\bm{x}_{1},\ldots,\bm{x}_{n}\in\mathbb{R}^{d}$ generating
	$D$. The EDM $D$ has a corresponding Gram matrix $M\in\mathbb{R}^{n\times n}$
	by the relationship
	\begin{equation}
	M_{ij}=\langle\bm{y}_{i},\bm{y}_{j}\rangle_{2}=\frac{1}{2}(D_{1j}+D_{i1}-D_{ij})
	\end{equation}
	with $\bm{y}_{k}=\bm{x}_{k}-\bm{x}_{1},\:k=1,\ldots,n$
	and vice versa
	\begin{equation}
	D_{ij}=M_{ii}+M_{jj}-2M_{ij}.\label{eq:M2D}
	\end{equation}
	
	The matrix $M$ furthermore has a specific structure
	\begin{equation}
	M=\left(\begin{array}{cc}
	0 & \bm{0}^{\top}\\
	\bm{0} & L
	\end{array}\right)\label{eq:L2M}
	\end{equation}
	with $L\in\mathbb{R}^{n-1\times n-1}$ and is symmetric and positive semi-definite.
	It therefore admits an eigenvalue decomposition $M=USU^{\top}=(U\sqrt{S})(U\sqrt{S})^{\top}=YY^{\top}$
	which, assuming that $S=\mathrm{diag}(\lambda_{1},\ldots,\lambda_{n})$
	with $\lambda_{1}\geq\lambda_{2}\geq\ldots\geq\lambda_{n}\geq0$,
	reveals a composition of coordinates $Y$ in the first $d$ rows where
	$d$ is the embedding dimension and the number of non-zero eigenvalues
	of $M$ associated to $D$.
	
	Therefore, the embedding dimension $d$ of $D$ is given by the rank of $M$ or equivalently the number of
	positive eigenvalues. In principle it would be sufficient to parameterize a symmetric positive semi-definite matrix
	$L\in\mathbb{R}^{n-1\times n-1}$, as it then automatically is also a Gram matrix for some set of vectors. However,
	also the set of symmetric positive semi-definite matrices behaves like a cone, which precludes the use of standard optimization techniques.
	
	Instead, we propose to parameterize an arbitrary symmetric matrix $\tilde{L}\in\mathbb{R}^{n-1\times n-1}$, as the set of
	symmetric matrices behaves like a vector space. This symmetric matrix can be transformed into a symmetric positive semi-definite matrix
	\begin{align}
	L = g(\tilde{L}) = g\left(U\begin{pmatrix}
	\lambda_1 & & \\ & \ddots & \\ & & \lambda_{n-1}
	\end{pmatrix}U^\top\right) = U\begin{pmatrix}
	g(\lambda_{1}) & & \\ & \ddots & \\ & & g(\lambda_{n-1})
	\end{pmatrix}U^\top
	\label{eq:make-spd}\end{align}
	by any non-negative function $g(\cdot)$ and then used to reconstruct $D$ via~(\ref{eq:M2D}) and~(\ref{eq:L2M}).
	
	This approach is shown in Algorithm~\ref{alg:edmgen} for the context of neural networks and the particular
	choice of $g = \mathrm{sp}$, the softplus activation function. A symmetric matrix $\tilde{L}$ is parameterized and transformed
	into a Gram matrix $M$ and a matrix $D$. For $M$ there is a loss in place that drives it towards a specific rank and for $D$
	we introduce a penalty on negative eigenvalues of~(\ref{eq:edmcondition}). It should be noted that $g(\cdot)$ can also be applied for the largest $d$ eigenvalues and explicitly set to $0$ otherwise, in which case the rank of $M$ is automatically $\leq d$. In that case it is not necessary to apply $L_\mathrm{rank}$.
	\begin{algorithm}
		\caption{Algorithm to train a generative neural network to (in general non-uniformly) sample Euclidean
			distance matrices based on the neural network $G$, where $N_{z}$ is the dimension of the input vector,
			$m$ the batch size, and $n$ the number of points to place relative
			to one another.\label{alg:edmgen}}
		
		\begin{algorithmic}[1]
			\State{Sample $\textbf{z}\sim\mathcal{N}(0,1)^{m\times N_z}$, i.e., sample from a simple prior distribution,}
			\State{Transform $X = G(\mathbf{z}) \in\mathbb{R}^{m\times (n-1)\times (n-1)}$ via a neural network $G$,}
			\For{$i=1$ to $m$}
			\State{Symmetrize $\tilde{L} \leftarrow \frac{1}{2}\left( X_i + X_i^\top \right)$}
			\State{Make positive semi-definite $L\leftarrow \mathrm{sp}(\tilde{L})$ with~(\ref{eq:make-spd})}
			\State{Assemble $M = M(L)$ with~(\ref{eq:L2M})}
			\State{Assemble $D = D(M)$ with~(\ref{eq:M2D})}
			\State{Compute eigenvalues $\mu_{1},\ldots,\mu_{n}$ of $-\frac{1}{2}JDJ$, see~(\ref{eq:edmcondition})}
			\State{$L_{\mathrm{edm}}^{(i)} \leftarrow \sum_{k=1}^n\mathrm{ReLU}(-\mu_k)^2$}
			\State{Compute eigenvalues $\lambda_{1},\ldots,\lambda_{n}$ of $M$ such that $\lambda_1\geq\lambda_2\geq\ldots\lambda_n$}
			\State{$L_\mathrm{rank}^{(i)} \leftarrow \sum_{k=d+1}^n \lambda_k^2$}
			\EndFor
			\State{$L \leftarrow \eta_1 \frac{1}{m}\sum_{i=1}^m L_\mathrm{edm}^{(i)} + \eta_2\frac{1}{m}\sum_{i=1}^m L_\mathrm{rank}^{(i)}$}
			\State{Optimize weights of $G$ with respect to $\nabla L$.}
		\end{algorithmic}
	\end{algorithm}

	\section{Euclidean distance matrix WGAN\label{sec:edm-wgan}}
	
	We consider the class of generative adversarial networks~(\cite{goodfellow2014generative})
	(GANs) and in particular Wasserstein GANs (WGANs), i.e., the ones that minimize the Wasserstein-1 distance in contrast to the original formulation, where the former can be related to minimizing the Jensen-Shannon
	divergence~\cite{arjovsky2017wasserstein}. WGANs consist
	of two networks, one generator network $G(\cdot)$, which transforms
	a prior distribution---in our case a vector of white noise $\boldsymbol{z}\sim\mathcal{N}(0,1)^{n_z}$---into a target distribution $\mathbb{P}_{g}$
	which should match the data's underlying distribution $\mathbb{P}_{r}$
	as closely as possible. The other network is a so-called critic network
	$C(\cdot)\in\mathbb{R}$, which assigns scalar values to individual
	samples from either distribution. High scalar values express that the sample is believed to belong to $\mathbb{P}_r$, low scalar values indicate $\mathbb{P}_g$. The overall optimization objective
	reads
	\begin{equation}
	\min_{G}\max_{C\in\mathcal{D}}\mathbb{E}_{\bm{x}\sim\mathbb{P}_{r}}\left[C(\bm{x})\right]-\mathbb{E}_{\bm{x}\sim\mathbb{P}_{g}}\left[C(\bm{x})\right],\label{eq:wgan}
	\end{equation}
	
	where $\mathcal{D}$ is the set of all Lipschitz continuous functions
	with a Lipschitz constant $L\leq1$. We enforce the Lipschitz constant
	using a gradient penalty (WGAN-GP) introduced in~\cite{gulrajani2017improved}.
	One can observe that the maximum in Eq.~(\ref{eq:wgan}) is attained
	when as large as possible values are assigned to samples from $\mathbb{P}_{r}$
	and as small as possible values to samples from $\mathbb{P}_{g}$. Meanwhile
	the minimum over the generator network $G$ tries to minimize that
	difference, which turns out to be exactly the Wasserstein-1 distance
	according to the Kantorovich--Rubinstein duality~(\cite{villani2008optimal}).
	Since the Wasserstein-1 distance is a proper metric of distributions,
	the generated distribution $\mathbb{P}_{g}$ is exactly the data distribution
	$\mathbb{P}_{r}$ if and only if the maximum in Eq.~(\ref{eq:wgan})
	is zero. The networks $G$ and $C$ are trained in an alternating
	fashion.
	
	We choose for the critic network the message-passing neural network
	SchNet~\cite{schutt2017quantum,schutt2017schnet,schutt2018schnet}
	$C_{\mathrm{SchNet}}(\cdot)$, which was originally designed to compute energies of molecules.
	
	It operates on the pairwise distances $(\sqrt{D_{ij}})_{i,j=1}^n$, $D\in\mathbb{EDM}^n$ in a structure
	and the atom types $\mathcal{T}^{n}$. If there is no atom type information present,
	these can be just constant vectors that initially carry no information.
	These atom types are then embedded into a state vector and transformed with variable sharing across
	all atoms. Furthermore there are layers in which continuous convolutions
	are performed based on the relative distances between the atoms. In a physical
	sense this corresponds to learning energy contributions of, e.g., bonds and angles.
	Finally all states are mapped to a scalar and then pooled in a sum.
	
	Due to the pooling and the use of only relative distances but never absolute coordinates,
	the output is invariant under translation, rotation, and permutation.
	
	The generator network $G$ employs the construction of Section~\ref{sec:edm} to
	produce approximately EDMs with a fixed embedding dimension.
	Therefore this architecture is able to learn distributions of Euclidean distance matrices.
	
	\section{Application and results}
	
	The WGAN-GP introduced in Sec.~\ref{sec:edm-wgan} is applied to
	a subset of the QM9 dataset
	consisting of $6095$ isomers with the chemical formula $\mathrm{C}_{7}\mathrm{O}_{2}\mathrm{H}_{10}$.
	To this end the distribution not only consists of the Euclidean distance matrices describing the
	molecular structure but also of the atom types. The generator produces an additional type
	vector in a multi-task fashion which is checked against a constant type reference with a cross-entropy loss.
	In particular this means that there is another linear transformation between the output of the neural network
	parameterizing the symmetric matrix $\tilde{L}$ (see, e.g.,~(\ref{eq:make-spd})) and a vector $\mathbf{t}\in\mathbb{R}^{m\times n\times n_\mathrm{types}}$ which is due to the use of a softmax non-linearity a
	probability distribution over the types per atom. This probability distribution is compared against a one-hot encoded type vector $\mathbf{t}_\mathrm{ref}$ representing the type composition in the considered isomer with a cross entropy term $H(\mathbf{t}, \mathbf{t}_\mathrm{ref})$. As in this example the chemical composition never changes, the type vector $\mathbf{t}_\mathrm{ref}$ is always constant.
	
	Furthermore the prior of a minimal distance between atoms is applied, i.e., we have a loss penalizing
	distances that are too small. Altogether we optimize the losses
	\begin{align}
	L_\mathrm{critic} &= \mathbb{E}_{(D,\textbf{t})\sim \mathbb{P}_g}\left[ C(D, \textbf{t})\right] - \mathbb{E}_{(D,\textbf{t})\sim \mathbb{P}_r}\left[ C(D, \textbf{t})\right] &&\text{(original WGAN loss)}\\
	&\quad + \lambda L_\mathrm{GP} &&\text{(gradient penalty of WGAN-GP)}\\
	&\quad + \varepsilon_\text{drift}\mathbb{E}_{(D, \textbf{t})\sim\mathbb{P}_r}\left[ C(D, \textbf{t})^2 \right] &&\text{(drift term~\cite{karras2017progressive})}\\
	L_\mathrm{gen} &= -\mathbb{E}_{(D,\textbf{t})\sim \mathbb{P}_g}\left[ C(D, \textbf{t})\right] &&\text{(original WGAN loss)}\\
	&\quad + \mathbb{E}_{(D,\textbf{t})\sim \mathbb{P}_g}\left[H(\textbf{t}, \textbf{t}_\mathrm{ref})\right] &&\text{(cross entropy for types)}\label{eq:Lgencrossentropy}\\
	&\quad + k\cdot \mathbb{E}_{(D,\textbf{t})\sim \mathbb{P}_g}\left[ \frac{1}{2}\sum_{i\neq j} (\sqrt{D_{ij}} - r)^2  \right] &&\text{(harmonic repulsion)}\\
	&\quad + L_\text{edm} &&\text{(for EDMs, see Alg.~\ref{alg:edmgen})}
	\end{align}
	with $C(\cdot)$ being a SchNet critic, $\textbf{t}_\mathrm{ref}$ a reference type order, $\lambda = 10$, $\varepsilon_\text{drift} = 10^{-3}$, $k=10$, and $r$ being the minimal pairwise distance achieved in the considered QM9 subset. The drift term leads to critic values around $0$ for real samples, as otherwise only the relative difference between values for real and fake samples matters. Although in principle the cross entropy loss~(\ref{eq:Lgencrossentropy}) is not required we found in our experiments that it qualitatively helps convergence. The generator network $G(\cdot)$ uses a combination of deconvolution and dense
	layers.
	
	The function $g(\cdot)$ ensuring positive semi-definiteness~(\ref{eq:make-spd}) was chosen to be the softplus activation $g = \mathrm{sp}$ for the largest three eigenvalues and we explicitly set all other eigenvalues to zero. This leads to a Gram matrix with exactly the right rank and the constraint does not need to be weakly enforced anymore in the generator's loss.
	
\begin{figure}
		\includegraphics[width=1\columnwidth]{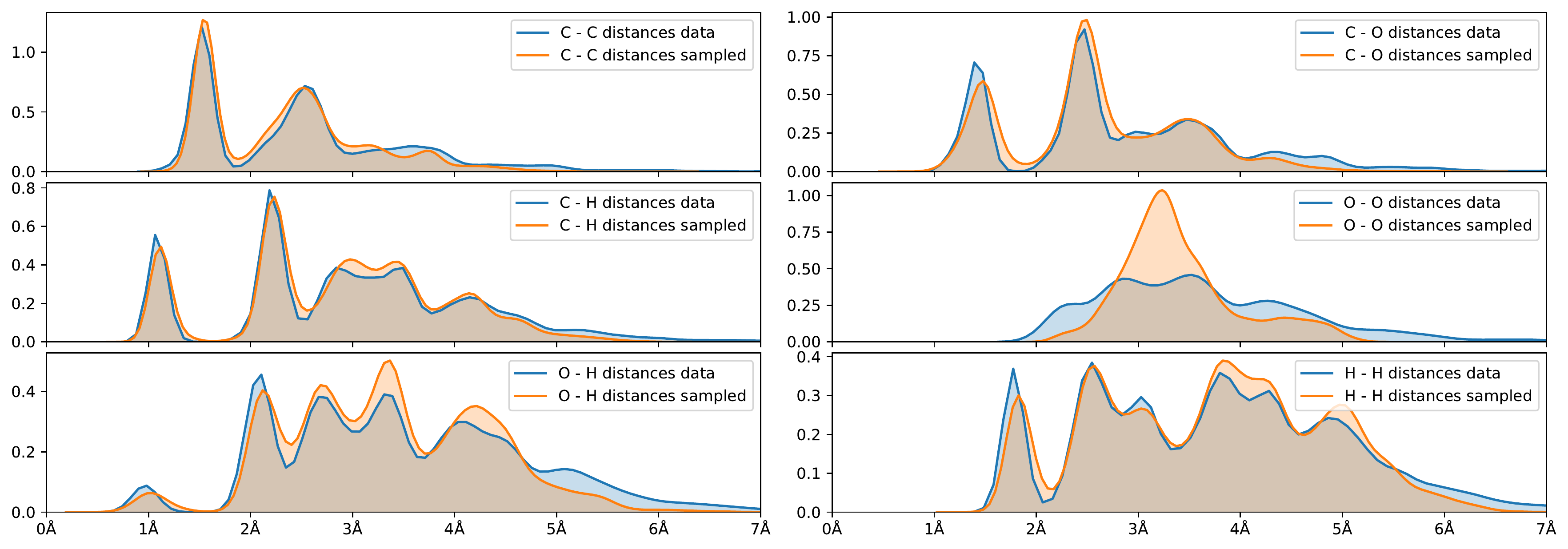}
		
		\caption{Distribution of pairwise distances between different kinds of atom
			type after training a Euclidean distance matrix WGAN-GP (Sec.~\ref{sec:edm-wgan})
			on the $\mathrm{C}_{7}\mathrm{O}_{2}\mathrm{H}_{10}$ isomer subset
			of QM9.\label{fig:distdist}}
	\end{figure}

	Prior to training the data was split into $50\%$ training and $50\%$ test data. After training on the training data set we evaluate the distribution of pairwise distances between different types of atoms, see Fig.~\ref{fig:distdist}. The overall shape of the distributions is picked up and only the distance between pairs of oxygen atoms are not completely correctly distributed.
	
	After generation we perform a computationally cheap validity test by inferring bonds and bond orders with Open Babel~\cite{o2011open}. On the inferred bonding graph we check for connectivity and valency, i.e., if for each atom the number of inferred bonds add up to its respective valency. This leaves us with roughly $7.5\%$ of the generated samples.
	
\begin{figure}
		\centering
		\includegraphics[width=.5\columnwidth]{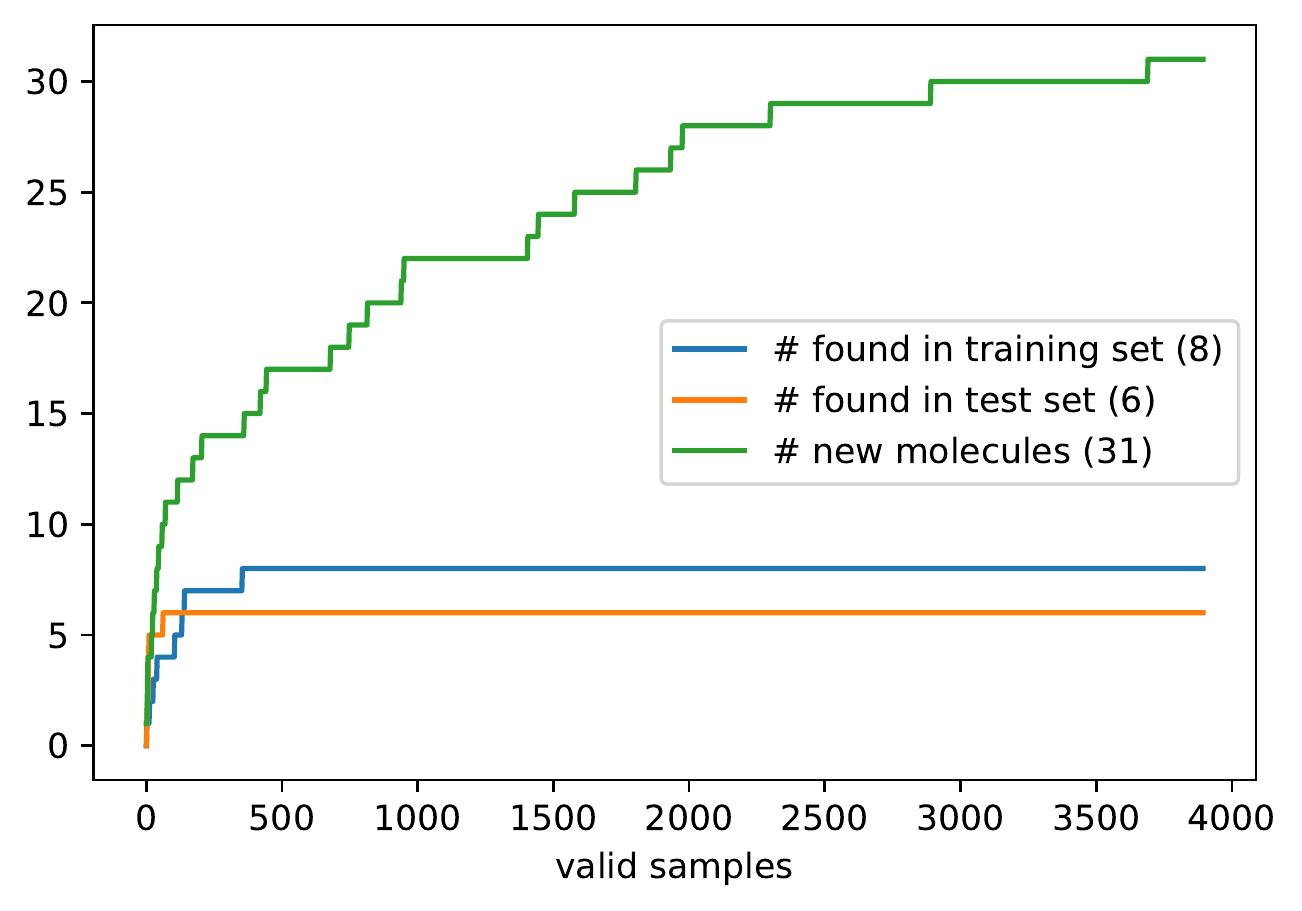}
		\caption{Number of unique molecular structures in terms of their topology for roughly $4000$ valid generated samples and whether they could be found in the training set (blue), the test set (orange), or had a new topology altogether (green).}
		\label{fig:smiles-molecules}
	\end{figure}

	For the valid samples we infer canonical SMILES representations which are a fingerprint of the molecule's topology in order to determine how many different molecule types can be produced using the trained generator. Fig.~\ref{fig:smiles-molecules} shows the cumulative number of unique SMILES fingerprints when producing roughly $4000$ valid samples. It can be observed that the network is able to generalize out of the training set and is able to generate not only topologies which can be found in the test set but also entirely new ones. Nevertheless, while the GAN implementation of our approach generates substantial diversity in configuration space, the diversity in chemical space is limited. This can be improved by better hyperparameter selection or choosing different generative structures, e.g.,~\cite{gebauer2018generating,gebauer2019symmetry}.
	
	While SMILES can be used to get an idea about the different bonding structures that were generated, it contains no information about different possible conformations in these bonding structures. To analyze the number of unique conformations that were generated, we compared each generated structure against all structures in the considered QM9 subset. Since the architecture is designed in such a way that it is permutation invariant, i.e., applying the critic onto a matrix $D=(D_{i,j})_{i,j=1}^n$ and $D_\sigma = (D_{\sigma(i), \sigma(j)})_{i,j=1}^n$ for some permutation $\sigma$ yields the same result, one first has to find the best possible assignment of atoms.
	
	To this end, we apply the Hungarian algorithm~\cite{Kuhn_Naval55_HungarianMethod} onto a cost matrix $C\in\mathbb{R}^{n\times n}$ for EDMs $D_1, D_2$ and type vectors $\textbf{t}_1, \textbf{t}_2\in\mathbb{R}^n$ with
	\begin{align}
	C_{i,j} = \begin{cases}
	\left|\frac{1}{n}\sum_{k=1}^n (D_1)_{i, k} - \frac{1}{n}\sum_{k=1}^n(D_2)_{j, k}\right| &\text{, if }(\textbf{t}_1)_i = (\textbf{t}_2)_j,\\
	\infty &\text{, otherwise.}
	\end{cases}
	\end{align}
	Intuitively this means that the cost of assigning atom $i$ in the first structure to atom $j$ in the second structure depends on whether their atom types match, in which case we compare the mean distance from the $i$-th atom to all other atoms in its structure to the same quantity for the $j$-th atom in the second structure. If the atom types do not match, we assign a very high number so that this particular mapping is not considered. After we have found an assignment between the atoms, we superpose the structures using functionality from the software package MDTraj~(\cite{McGibbon_BJ15_MDTraj}) and evaluate the maximal atomic distance between all heavy atoms (i.e., carbons and oxygens) after alignment. The cutoff at which we consider a structure to be a distinct conformation is a maximal atomic distance between heavy atoms of more than $d_\mathrm{cutoff} = \SI{0.6}{\angstrom}$, i.e., more than half a carbon--carbon bond length.
	
\begin{figure}
		\centering
		\includegraphics[width=.5\columnwidth]{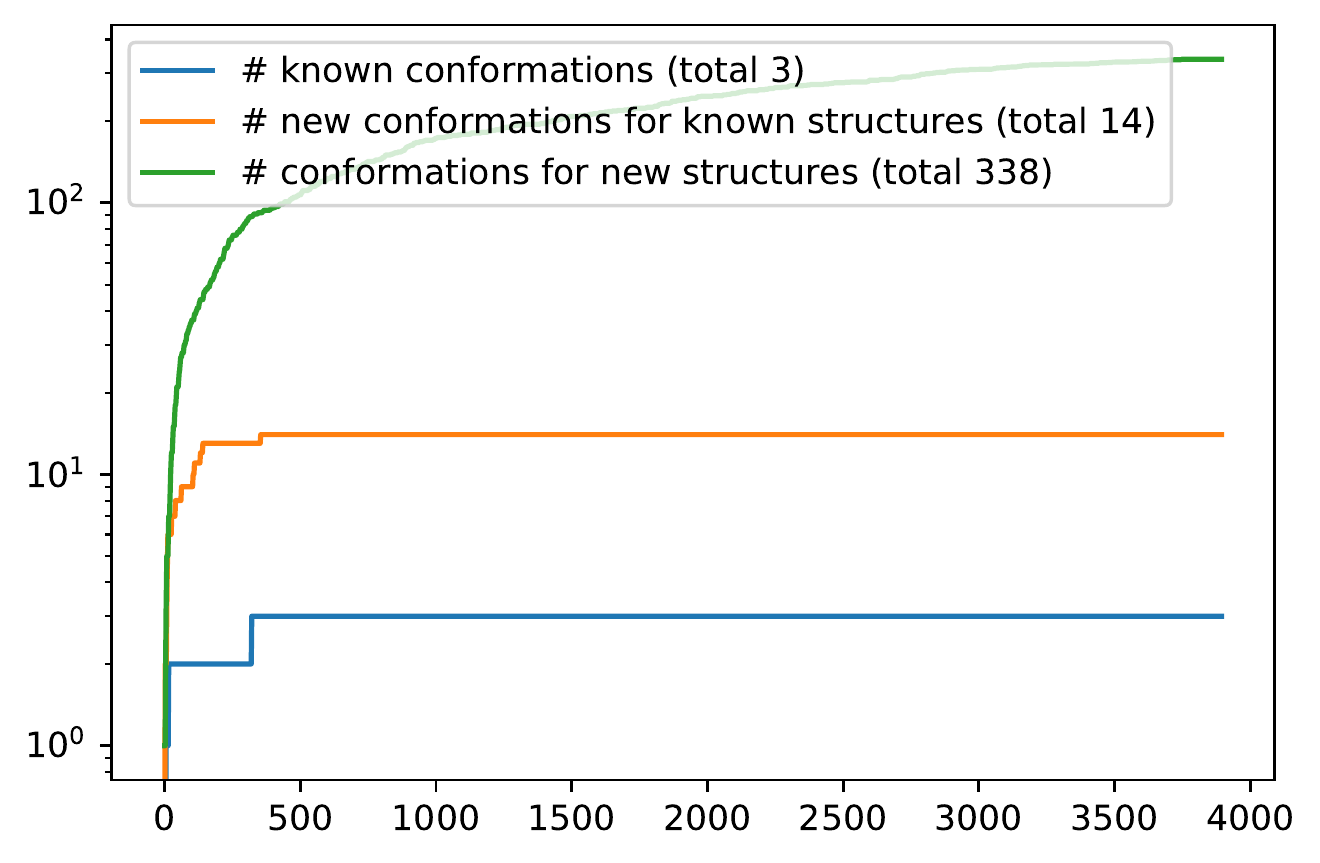}
		\caption{Unique generated conformations up to a maximal heavy atom distance cutoff of $d_\mathrm{cutoff} = \SI{0.6}{\angstrom}$ after assignment and superposition. We distinguish the categories of known conformations in the considered subset of the QM9 database (blue), new conformations for contained molecular structures (orange), and distinct conformations for molecular structures that are not contained.}
		\label{fig:conformations}
	\end{figure}

	The results of this analysis are depicted in Fig.~\ref{fig:conformations}. One can observe that although the reported number of unique molecular structures via SMILES is rather low, under our similarity criterion a lot of different valid conformations are discovered; in particular also some new conformations of structures that were already contained in the QM9 database.
	
\begin{figure}
		\centering
		\includegraphics[width=.7\columnwidth]{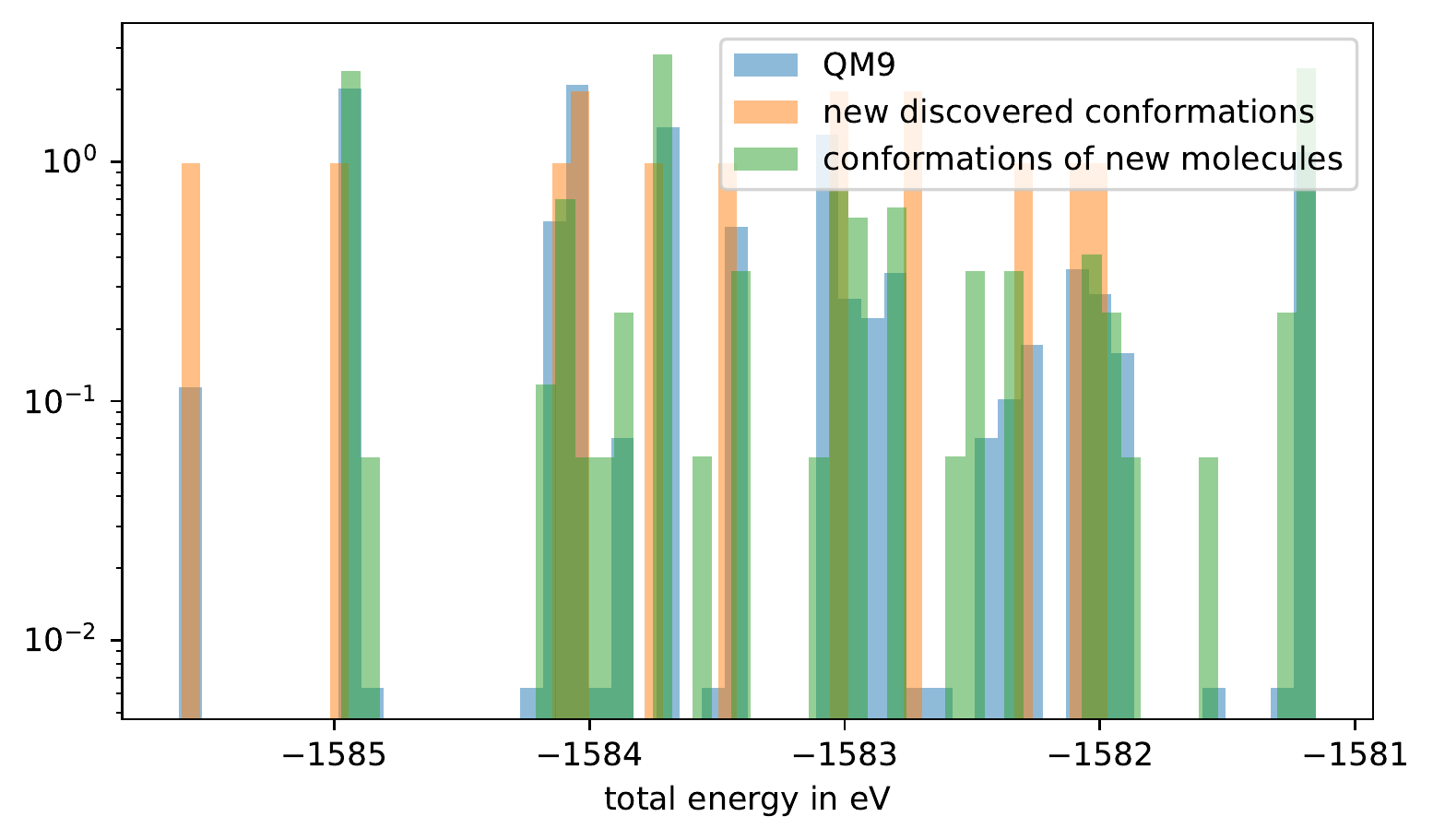}
		\caption{Total energies of structures that were relaxed with the semi-empirical method implemented by MOPAC, in particular for molecules contained in the considered QM9 subset (blue), structures that correspond to new conformations for contained molecules (orange), and unique conformations that belong to new molecules.}
		\label{fig:energies}
	\end{figure}

	Finally, we also check for the approximate total energies of the generated molecules compared to the database's. To this end, we use the semi-empirical method provided by the software package MOPAC~\cite{stewart1990mopac} to relax all structures in the considered QM9 subset as well as all valid generated structures, see Fig.~\ref{fig:energies}. It can be observed that after relaxation all energies are contained within the same range of roughly $\SI{-1586}{\electronvolt}$ to $\SI{-1581}{\electronvolt}$. This means that our architecture is capable to propose structures which after relaxation have energies comparable to the ones in the database.
	
\begin{figure}
		\centering
		\includegraphics[width=\columnwidth]{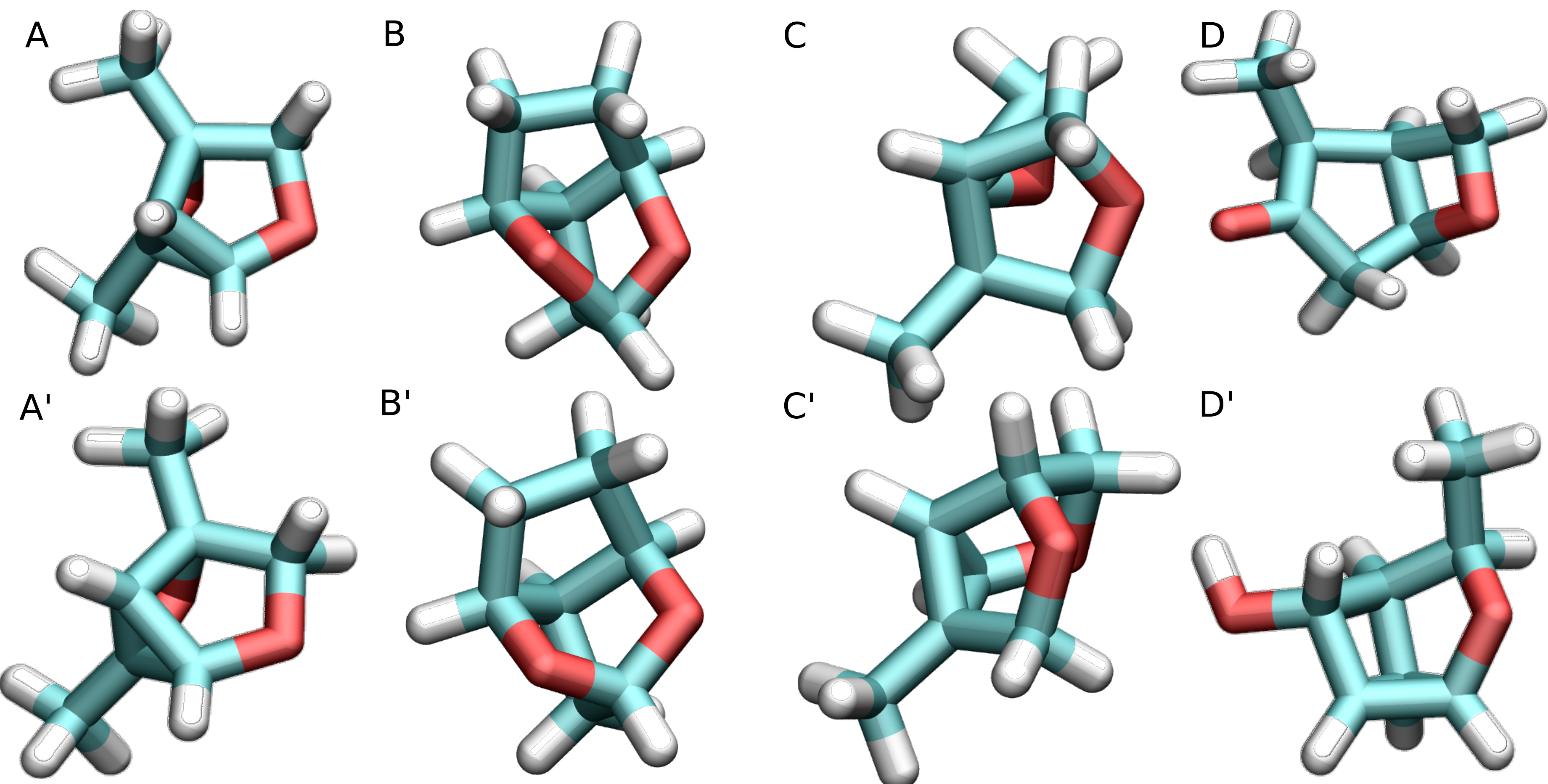}
		\caption{Sampled structures with the Euclidean distance matrix WGAN. Top row A to D are generated samples, bottom row A' to D' are closest matches from the QM9 database. Generated molecules A and B could be matched with A' and B' up to a maximum atom distance of $\SI{0.6}{\angstrom}$. Generated molecules C and D are new molecular structures with their closest matches C' and D', respectively.}
		\label{fig:samples}
	\end{figure}

	In Figure~\ref{fig:samples} we show examples of generated molecules in the top row (A--D) and the closest respective matches in the QM9 database in the bottom row (A'--D'). The closeness of a match was determined by the maximal atomic distance after assignment of atom identities and superposition. Configurations A and B could be matched with a maximal atomic distance of less than $d_\mathrm{cutoff}$.
	
	\section{Conclusion and discussion}
	We have developed a way to parameterize the output of a neural network so that it produces valid Euclidean distance matrices with a predefined embedding dimension without placing coordinates in Cartesian space directly. This enables us to be naturally invariant under rotation and translation of the described object.
	Given a network that is able to produce valid Euclidean distance matrices we introduce a Wasserstein GAN that can learn to one-shot generate distributions of point clouds irrespective of their orientation, translation, or permutation. The permutation invariance is achieved by incorporating the message passing neural network SchNet as critic.
	
	We applied the introduced WGAN to the $\mathrm{C}_7\mathrm{O}_2\mathrm{H}_{10}$ isomer subset of the QM9 molecules database and could generalize out of the training set as well as achieve a good representation of the distribution of pairwise distances in this set of molecules.
	
	In future work we want to improve on the performance of the network on the isomer subset as well as extend it to molecules of varying size and chemical composition. We expect the ideas of this work to be applicable for, e.g., generating, transforming, coarse graining, or upsampling point clouds.
	
	To improve the quality of the generated molecular structures a follow up of this work would be including a force field so that more energetically reasonable configurations are produced and including penalty terms which favor configurations that do not contain invalid valencies, i.e., produce valid molecular structures. Furthermore optimizing a conditional distribution for a particular molecule---i.e., conditioning on the molecular graph---and exploring its conformations is a natural extension of this work.
	
	\subsubsection*{Acknowledgments}
	We acknowledge funding from European Commission (ERC CoG 772230 “ScaleCell”), Deutsche Forschungsgemeinschaft (CRC1114/A04), the MATH+ Berlin Mathematics research center (AA1x6, EF1x2).  We are grateful for inspiring discussions with Tim Hempel, Brooke Husic, Simon Olsson, Jan Hermann, and Mohsen Sadeghi (FU Berlin).
	
	\bibliography{literature.bib}
	\bibliographystyle{unsrt}
\end{document}